\pdfoutput=1
\documentclass[11pt]{article}

\usepackage{ACL2023}

\usepackage{times}
\usepackage{latexsym}

\usepackage[T1]{fontenc}
\usepackage[utf8]{inputenc}

\usepackage{microtype}

\usepackage{inconsolata}

\usepackage{graphicx}
\usepackage{multirow}
\usepackage{footnote}
\usepackage{footmisc}
\usepackage{subfigure} 
\usepackage{url}
\usepackage{booktabs}
\usepackage{linguex}
\usepackage{booktabs} 
\usepackage{amsmath}
\usepackage{fontawesome}

\AtBeginDocument{%
  \settowidth{\Exlabelwidth}{(0)}%
}
\title{GUMSum: Multi-Genre Data and Evaluation \\ for English Abstractive Summarization} % Out of Domain 

\author{Yang Janet Liu \and Amir Zeldes \\
        Department of Linguistics \\ Georgetown University \\ 
        {\tt \{yl879, amir.zeldes\}@georgetown.edu}}
\begin{document}
\maketitle

\begin{abstract}

Automatic summarization with pre-trained language models has led to impressively fluent results, but is prone to `hallucinations', low performance on non-news genres, and outputs which are not exactly summaries. Targeting ACL 2023's `Reality Check' theme, we present GUMSum, a small but carefully crafted dataset of English summaries in 12 written and spoken genres for evaluation of abstractive summarization. 
Summaries are highly constrained, focusing on substitutive potential, factuality, and faithfulness. We present guidelines and evaluate human agreement as well as subjective judgments on recent system outputs, comparing general-domain untuned approaches, a fine-tuned one, and a prompt-based approach, to human performance. 
Results show that while GPT3 achieves impressive scores, it still underperforms humans, with varying quality across genres. Human judgments reveal different types of errors in supervised, prompted, and human-generated summaries, shedding light on the challenges of producing a good summary. 
\end{abstract}

\section{Introduction}
\label{sec:intro}

Recent advances in supervised summarization models as well as prompt-based approaches using large pre-trained language models have led to substantial improvements in summary fluency, with prompt-based outputs now surpassing supervised approaches in human evaluation \cite{NewsSumGPT3}. At the same time, researchers in the field repeatedly note that the most commonly used datasets, such as CNN/DailyMail (CNN/DM, \citealt{CNNDM}) and Extreme Summarization (XSum, \citealt{narayan-etal-2018-dont}), which are large-scale `found' datasets not designed to facilitate high quality summarization, are problematic, and in many cases contain texts which are not summaries, are incomplete or unfaithful to the texts they relate to, add information not present in texts, or any combination of the above \cite{ehudreiter2022post,RevisitingGoldStandard4Summarization}. 
Existing datasets are also limited to mainly newswire text (cf.~\citealt{zopf-etal-2016-next}), which is a fraction of extant genres in general and on the Web.

The main contributions of this paper are in providing and evaluating a very genre-diverse dataset and guidelines for carefully crafted, rather than `found' summaries, which follow the same design across text types. Building on the UD English GUM treebank \cite{Zeldes2017}, which contains 213 spoken and written texts balanced across 12 different genres, our summaries target three goals: 1) to be \textbf{substitutive} (i.e.~informative, functioning as a substitute for reading a text, cf.~\citealt{Edmundson1969AutomaticExtracting,Nenkova2011McKeown}) rather than indicative (e.g.~`clickbait' designed to attract readership); 2) to be \textbf{faithful} to the text, adhering to original formulations wherever possible; 3) to be \textbf{hallucination-free}, meaning summaries make a strong effort not to add any information (even if it is likely to be true), mentioning only entities and events actually contained in the text, thereby preventing typical errors associated with datasets such as XSum \cite{narayan-etal-2018-dont}. 
Instructions on obtaining the dataset and responses from the human evaluation study as well as evaluation code can be found at \url{https://github.com/janetlauyeung/GUMSum4EVAL}.\footnote{Data is also available from the corpus website at \url{https://gucorpling.org/gum/} and guidelines at \url{https://wiki.gucorpling.org/en/gum/summarization}. }

\section{Related Work}
\label{sec:related-work}

The problem of mitigating factuality and faithfulness issues in Natural Language Generation (NLG) has recently received considerable attention, with studies proposing auxiliary tasks using the Multi-Task Learning approach to constrain models, such as overlapping entities \cite{nan-etal-2021-entity}, encoding of predicate triples from source documents \cite{zhu-etal-2021-enhancing} or encouraging systems to incorporate or copy entities from source documents \cite{XiaoCarenini2022,maddela-etal-2022-entsum}. In addition, \citet{UnderstandingFactualErrorsinSum2022Durrett} present a thorough investigation of factual errors in summarization and propose a taxonomy of error types  with a focus on entity and predication errors, while \citet{THOMSON2023101482} examine types of accuracy errors made by neural systems and contrast them with human errors. 

These papers share concerns about the nature of widely used datasets for English, such as XSum and CNN/DM, but are limited by the lack of evaluation data specifically targeting genre-diverse texts with high-quality summaries: ones which ideally maximize faithfulness, rule out hallucinations, and follow consistent guidelines for what constitutes a summary. Although there are some non-news single-document summarization datasets covering Reddit \cite{kim-etal-2019-abstractive} and Podcast data \cite{PodcastSummary2022SIGIR}, text types are still quite limited and data is often not publicly available \cite{UnderstandingFactualErrorsinSum2022Durrett}. This motivates our work to create open-access, multi-genre data with consistent guidelines across text types.

\section{Dataset}
\label{sec:dataset}

\paragraph{Contents}
GUMSum covers the 213 documents (amounting to $\sim$200K tokens) from the 12-genre UD English GUM corpus (\citealt{Zeldes2017}; specifically GUM V9), which provides gold syntax trees, entity types, coreference resolution, and discourse parses for the data. For this paper, we added summaries to each document in the corpus, by the authors and students in a Computational Linguistics course as part of a class-based project,\footnote{Consent to release data was given by all students.} guided by general and genre-specific instructions. Although the range of $\sim$20 human summarizers is broad as a result, we defined guidelines to constrain summaries and ensure they are maximally `reality-checked', i.e.~faithful and factual, as evaluated below. Documents vary in length, ranging between 167 and 1,878 tokens (\texttt{mean}=957, \texttt{sd}=249.6), and cover the genres in Table \ref{tab:gum}. 
Because of the classroom context in which summaries are collected and the natural variation in student styles and adherence to guidelines, all summaries are thoroughly checked by a teaching assistant and the course instructor. For the 24 documents in the UD treebank's official \texttt{test} set of GUM V9, we provide two summaries to support inter-annotator agreement and multiple-reference evaluation.

\begin{table}[htbp]
\centering
\resizebox{\columnwidth}{!}{%
\begin{tabular}{@{}llccr@{}}
\toprule
\textbf{Genres} & \textbf{Source} & \textbf{Docs} & \textbf{Toks} & \textbf{{\o}Sum.Len (sd)} \\ \midrule
Interviews & Wikinews & 19 & 18,190 & 49 (6.3) \\
News stories & Wikinews & 23 & 16,145 & 51 (9.0) \\
Travel guides & Wikivoyage & 18 & 16,514 & 59 (8.9) \\
How-to guides & WikiHow & 19 & 17,081 & 67 (6.5) \\
Academic & various & 18 & 17,169 & 35 (11.2) \\
Biographies & Wikipedia & 20 & 18,213 & 44 (9.8) \\
Fiction & various & 19 & 17,510 & 47 (10.3) \\
Web forums & Reddit & 18 & 16,364 & 50 (8.7) \\
Conversations & SBC & 14 & 16,416 & 41 (13.7) \\
Speeches & various & 15 & 16,720 & 46 (9.2) \\
Vlogs & YouTube & 15 & 16,864 & 50 (11.8) \\
Textbooks & OpenStax & 15 & 16,693 & 51 (8.9) \\
\hline
\multicolumn{2}{l}{total / average} & 213 & 203,879 & 50 (12.2) \\
\bottomrule
\end{tabular}%
}
% \vspace{-6pt}
\caption{Overview and Statistics of GUMSum. } \label{tab:gum} % ($*$ = avg.~summary length by tokens).
\vspace{-12pt}
\end{table}

\paragraph{Guidelines} 
Previous literature has characterized `good' summaries primarily as ones that are concise, accurate, fluent, and coherent \cite{fabbri-etal-2021-summeval}. What these qualities mean varies depending on the summary's objective: whether it is domain-specific or general, indicative (enticing readers to read the text) or informative (aiming to substitute reading it, \citealt{Nenkova2011McKeown}) etc.~GUMSum's summaries explicitly target a \textbf{domain-general, substitutive, maximally concise} format, which is therefore constrained to:

\vspace{-3pt}
\begin{enumerate}
    \setlength\itemsep{0.2em}
    \item have at most one sentence / 380 characters\footnote{We follow XSum in targeting 1-sentence summaries, and we aimed for a maximum of 5 lines in a PEP8-compliant IDE, but in practice no summary exceeded 380 characters. }
    \item have the goal of replacing reading the text
    \item give participants/time/place/manner of events
    \item form a sentence rather than a fragment
    \item omit distracting information
    \item avoid entities or information not present in the text, even if we are fairly sure it is true
    \item reject synonyms for words in the text
\end{enumerate}
\vspace{-3pt}

For instance, the summary in \ref{ex:bank1} for a story involving `robbers plundering a vault' follows guidelines by providing a declarative-sentence (criteria \texttt{[1]}, \texttt{[4]}), synopsis of events, participants (exactly \textit{five robbers}), time (a date) and place (\textit{Poughkeepsie}) (\texttt{[3]}), as well as additional details (exact name of the bank, mode of escape). \ref{ex:bank2} is underspecified (we do not know when or where the event occurred, criterion \texttt{[3]}). \ref{ex:bank3} paraphrases the robbers' escape by introducing an entity \textbf{\textit{not}} in the original text (uncaught by \textit{police}, violating \texttt{[6]}), and substitutes `robbed' for `plundered', a near synonym but a deviation from the original text's style (\texttt{[7]}).  % criterion

\ex. \faThumbsOUp  \textit{On March 23, 1999, five bank robbers plundered the vault of First National Bank in Poughkeepsie, NY and escaped in a bus they had stolen.}\label{ex:bank1} 

\ex. \faThumbsODown  \textit{Bank robbers plundered a vault and escaped.}\label{ex:bank2} 

\ex. \faThumbsODown  \textit{Bank robbers who robbed a bank in Poughkeepsie were never caught by police.} \label{ex:bank3}

Although these examples illustrate newswire language, GUMSum covers very different spoken and written text types as well:

\ex. \textit{Some people debate whether the original 3 hour cut of Snyder's movie about Batman and Superman should have been released instead of the shorter version, which prioritized getting to the action faster in order to appeal to a general audience.} (Reddit)\label{ex:reddit}

\ex. \textit{Ash tells about her day, which includes a yoga class, marketing brand management class, doing some work while having coffee at Saxby's, and finally cooking pasta with peppers for dinner together with her boyfriend Harry.} (YouTube CC-BY vlog)\label{ex:vlog}

The summary in \ref{ex:reddit} follows the guidelines by not mentioning that the discussion is on Reddit (\texttt{[6]}, the interlocutors are simply `people'), since Reddit is not mentioned. Similarly, while Zack Snyder's film \textit{Batman v Superman: Dawn of Justice} is most likely being discussed, it is not named explicitly, leading to the formulation `Snyder's movie about Batman and Superman'. In \ref{ex:vlog}, the summary focuses on conveying events which happen over the course of a vlog, but again, the unmentioned word `vlog' is avoided, while specific details about the participants and circumstances (people, but also the type of class) are prioritized. Summaries are thus highly constrained to remain faithful and avoid even minor potential hallucinations, such as completing the title of a film. For more on genre-specific guidelines and examples, see Appendix \ref{appendix:guidelines}. 

\begin{figure*}[htp]
\centering
  \subfigure[human preferences]{\includegraphics[width=0.32\textwidth]{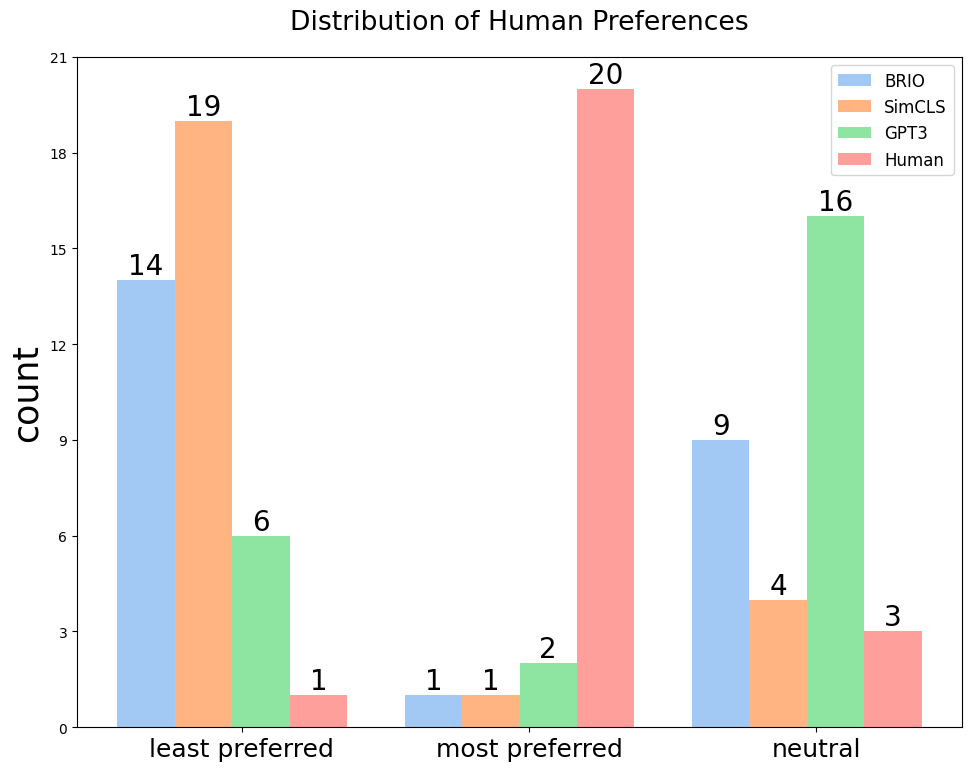}\label{fig:preference}}
  \subfigure[substitutive potential]{\includegraphics[width=0.32\textwidth]{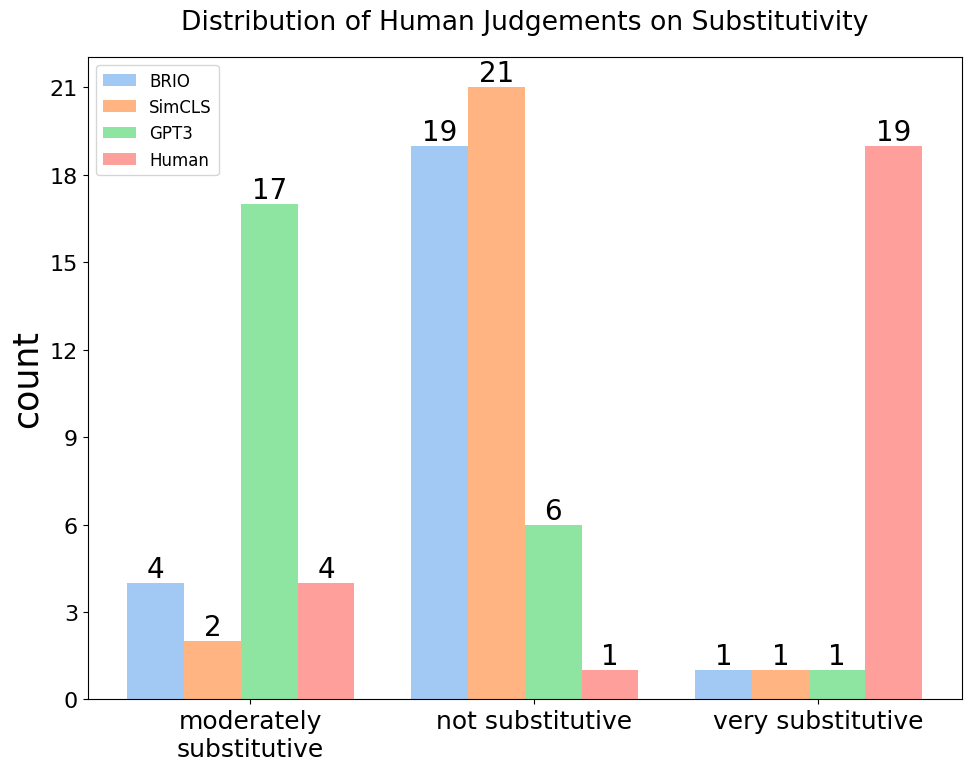}\label{fig:substitutive}}
  \subfigure[hallucination \& faithfulness]{\includegraphics[width=0.33\textwidth]{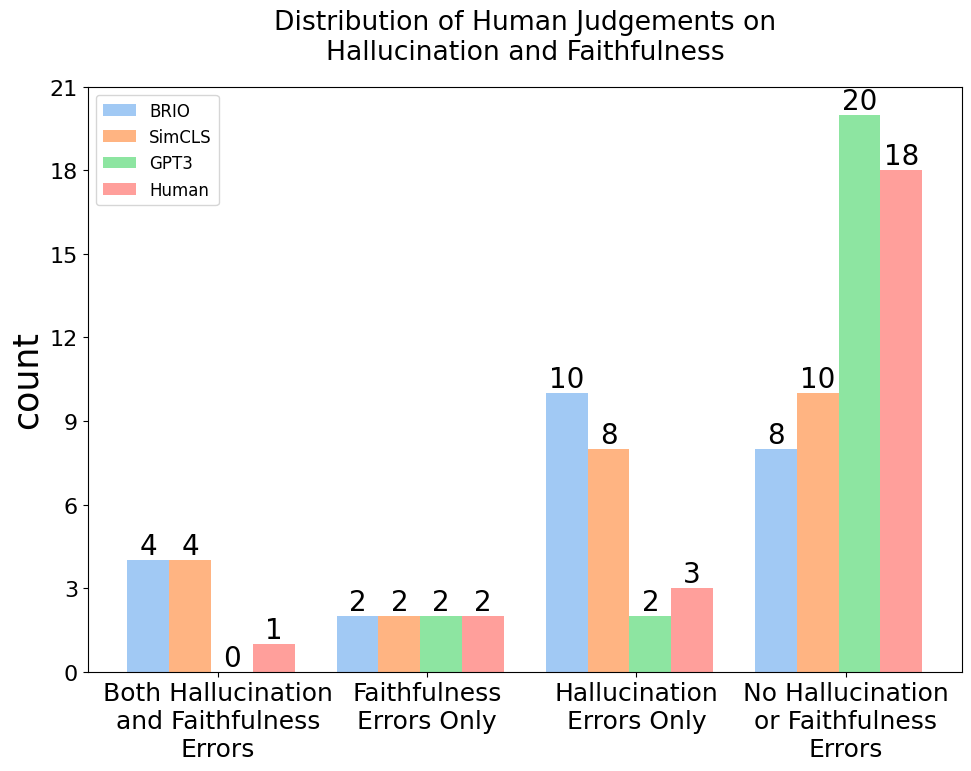}\label{fig:factuality}}
  % \subfigure[faithfulness]{\includegraphics[width=0.32\textwidth]{barplots/faithfulness.png}\label{fig:faithfulness}}
  % \subfigure[english]{\includegraphics[width=0.32\textwidth]{barplots/english.png}\label{fig:english-quality}}
  % \subfigure[source]{\includegraphics[width=0.32\textwidth]{barplots/source.png}\label{fig:source}}
    \vspace{-8pt}
  \caption{Barplots of Human Evaluations on Preferences, Substitutive Potential, Hallucination, and Faithfulness. }
  \label{fig:bar-plots}
  \vspace{-12pt}
\end{figure*}

\section{Evaluation}
\label{sec:evalution}

\paragraph{Automatic Evaluation} To evaluate how well current neural approaches produce `reality-checked' summaries approaching the ones in GUMSum, we obtain system outputs from two recent supervised systems, BRIO \cite{liu-etal-2022-brio} and SimCLS \cite{liu-liu-2021-simcls}, as well as prompt-based outputs using a GPT3 model \cite{GPT3BrownEtAal2020}, \texttt{GPT3-text-davinci-002} (GPT3-DV2), with the prompt `\textit{Summarize the text above in one sentence.}'. We chose system models trained on the XSum dataset, since it has one-sentence summaries more in line with the GUMSum data. However, because systems have never seen data in many of GUMSum's genres, we also add an additional experiment in which we fine-tune the higher-scoring supervised system, i.e.~BRIO's trained-model on XSum for \textit{generation},
% \footnote{\url{https://huggingface.co/Yale-LILY/brio-xsum-cased}}
by continuing training it on the 165 documents in the UD treebank's \texttt{train} set of the underlying GUM V9 corpus (\textsc{BRIO-FT} in Table \ref{tab:gum-test-scores}; details/splits and system output selection can be found in Appendix \ref{appendix:experiment-details}). Scores are compared to a second human-written summary obtained from a human evaluation study, using the same guidelines.

\begin{table}[ht]
\centering
\resizebox{\columnwidth}{!}{%
\begin{tabular}{@{}l|cccccccc@{}}
\toprule
 & \textbf{R-1} & \textbf{R-2} & \textbf{R-L} & \textbf{BS} & \textbf{MS} & \textbf{METEOR} & \textbf{BLEU} & \textbf{BLEURT} \\ \midrule
\textbf{SimCLS} & 23.1 & 6.2 & 17.2 & 86.0 & 12.1 & 13.4 & 2.1 & 31.9 \\
\textbf{BRIO} & 27.8 & 10.2 & 21.2 & 87.2 & 15.9 & 18.0 & 3.7 & 36.3 \\
\textbf{GPT3-DV2} & 31.1 & \textbf{12.1} & 25.1 & 88.5 & 21.1 & 20.8 & 3.8 & 42.2 \\ 
\textbf{BRIO-FT$^{*}$} & \textbf{37.3} & 12.0 & \textbf{27.1} & \textbf{88.7} & \textbf{27.4} & \textbf{27.6} & \textbf{6.1} & \textbf{44.3} \\ \midrule
\textbf{Human 2} & {\color[HTML]{0000FF} \textbf{38.9}} & {\color[HTML]{0000FF} \textbf{12.7}} & {\color[HTML]{0000FF} \textbf{28.4}} & {\color[HTML]{0000FF} \textbf{88.8}} & {\color[HTML]{0000FF} \textbf{28.5}} & {\color[HTML]{0000FF} \textbf{33.0}} & {\color[HTML]{0000FF} \textbf{7.5}} & {\color[HTML]{0000FF} \textbf{50.2}} \\ \bottomrule
\end{tabular}%
}
% \vspace{-3pt}
\caption{Automatic Evaluation Metrics of System Outputs and Human Agreement ($^{*}$ = 3 run average).}
\label{tab:gum-test-scores}
\vspace{-7pt}
\end{table}

\begin{figure}[htp]
 \centering
 \includegraphics[width=\columnwidth, scale=1]{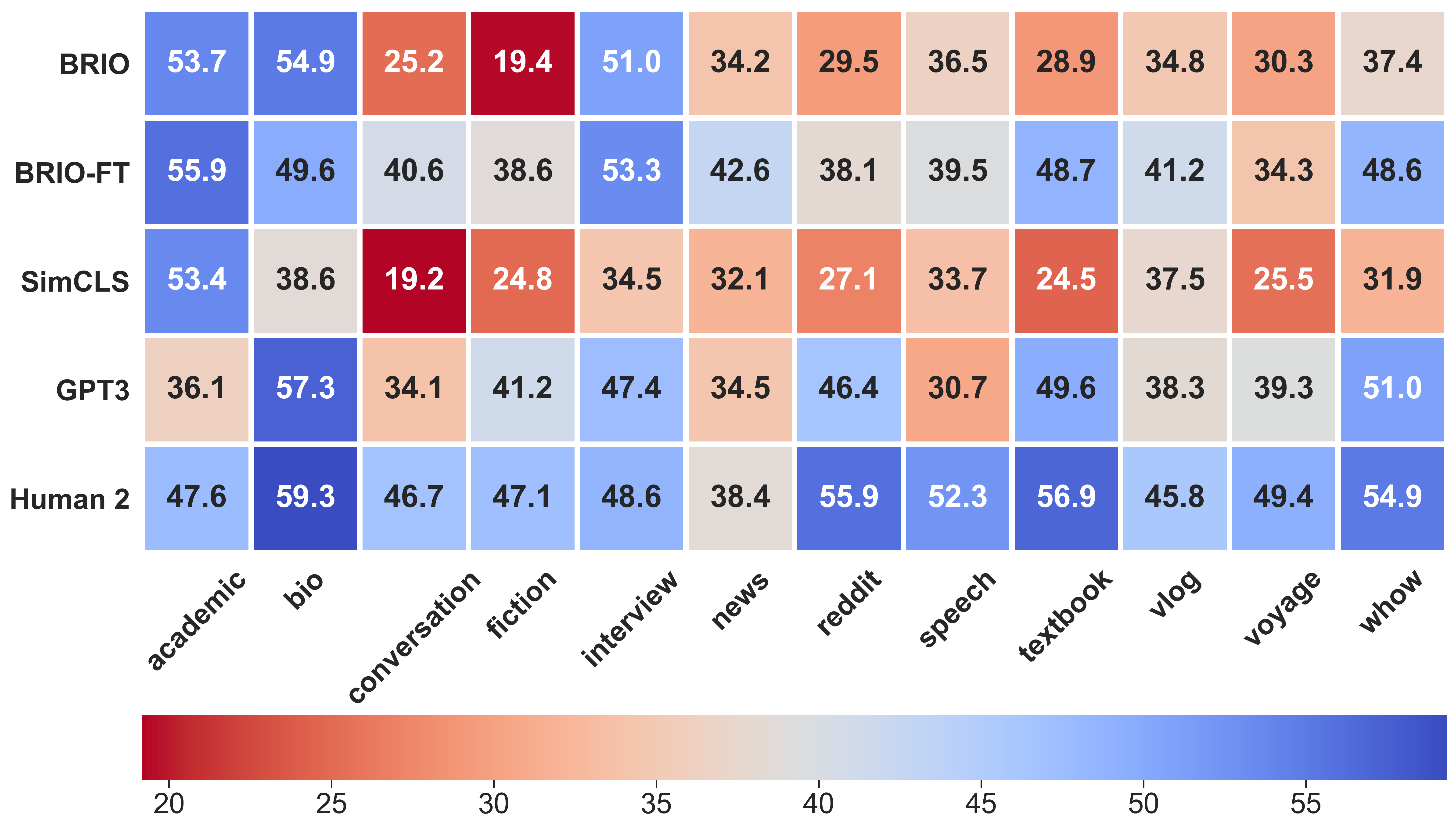}
 \caption{BLEURT Scores by Genre for Each Scenario (whow=how-to guides, voyage=travel guides).}
 \label{fig:genre-bleurt}
 % \vspace{-5pt}
\end{figure}

Table \ref{tab:gum-test-scores} shows that while systems have impressive scores for ROUGE \cite{lin-2004-rouge}, BERTScore (BS, \citealt{Zhang2020BERTScore}), MoverScore (MS, \citealt{zhao-etal-2019-moverscore}), METEOR \cite{banerjee-lavie-2005-meteor}, BLEURT \cite{sellam-etal-2020-bleurt}, and BLEU \cite{BLEU2002}, they still lag behind the human summaries across the board. Reproducing findings by \citet{NewsSumGPT3}, GPT3-DV2 outperforms supervised systems trained on XSum, though our data contains much more diverse genres than those in that paper. However, fine-tuning on even a small amount of GUMSum data (165 documents) in this paper already outperforms GPT3-DV2. This strongly suggests that a major problem with supervised systems in domain-general settings is simply the training data itself. Qualitative inspection of outputs suggests fine-tuning was particularly helpful for summarizing conversations, Reddit, and how-to guides, on which all systems struggled. For humans, genre differences were much less pronounced, with lowest scores surprisingly for news. 
Figure \ref{fig:genre-bleurt} gives a detailed breakdown of BLEURT scores \cite{sellam-etal-2020-bleurt} by genre for each scenario. Human scores lead in every genre except academic, news, and interview, and generally vary less by genre than systems. BRIO-FT is improved especially on genres that diverge from XSum, such as conversations, travel guides from Wikivoyage, and how-to guides from Wikihow.

Finally, the human scores provide some numbers for ceiling performance as reflected by automatic metrics. Comparing human numbers to the best-system numbers suggests that there is a substantial gap for systems which have never been trained on in-domain data. However, for the the fine-tuning (FT) scenario, we notice that ROUGE scores are neck-and-neck with the second human summary, likely because the system is trained with an objective averaging R1, R2, and R-L, on which it excels. By contrast, metrics more focused on verbatim overlap, such as BLEU, or semantic similarity, such as BLEURT, retain a more substantial gap, with FT results on BLEURT being close to GPT3-DV2 and still nearly 6 points below human performance.

It is an established finding however that metrics do not tell the whole story \cite{novikova-etal-2017-need,reiter2018structured,MarasovicGradient2018NLP,Gehrmann2022ObstaclesInNLGEval}. In fact, we regularly observe hallucinations, especially in XSum-trained systems, such as prefixing generic leads (e.g.~`\textit{In our series of letters from British journalists ...}', when there are no journalists involved) or inserting entities and events not mentioned in the text. We thus conduct a human evaluation of system outputs below, focusing on substituitivity, hallucinations, and faithfulness, and more importantly, apply the same evaluation criteria to the human-written summaries for a more targeted evaluation, as advocated by \citet{RevisitingGoldStandard4Summarization}.

\paragraph{Human Evaluation}
\label{sec:human-eval}
We asked 12 Linguistics students to evaluate the full texts and the summaries of the 24 documents in the \texttt{test} set of the source GUM V9 corpus and to produce an additional summary for their assigned texts (see detailed instructions in Appendix \ref{appendix:humaneval-details}).\footnote{The hourly pay is \$20.29/hour based on the pay rate of the 2022 / 2023 academic year for graduate students at Georgetown University. It took about 1.5 hours in total for each annotator to complete all the tasks for the two documents.}
Figure \ref{fig:bar-plots} shows humans overwhelmingly preferred the human-written summary (\ref{fig:preference}, 83\%, with exceptions citing gold summaries as less pleasant to read), and also found it best at substituting reading the text (\ref{fig:substitutive}, 79\%). Pretrained supervised systems were judged to be highly non-substitutive (88\% for SimCLS, 79\% for BRIO), while 71\% of GPT3-DV2 outputs were judged moderately so.

% avg. character length 
% BRIO: 125.41666666666667
% SimCLS: 113.04166666666667
% GPT3: 137.83333333333334 
% Human1 ("gold"): 272.3333333333333
% Human2: 298.2083333333333
While all systems exhibited some hallucinations and unfaithfulness, GPT3-DV2 performed best, in part because its outputs tended to be short (mean 138 characters vs.~human 272 characters) and general, giving fewer chances for issues. At the same time, hallucination types varied substantially. Human violations in both categories were rare and subtle, resulting from evaluators adhering to guidelines very literally: for example, one evaluator proposed that a human summary's use of the pronoun `she' in reference to a vlogger whose pronouns had not been stated is a form of hallucination, while another pointed out that a mention of `Washington' in a news article was a faithfulness issue, since without specifying `DC', the place is ambiguous. Hallucinations from GPT3-DV2 were more pronounced (e.g.~designating a speaker mentioning retirement as an attendee of a seminar about retirement, which was not mentioned), while XSum-trained systems had more extreme cases, such as incorrectly attributing a speech about New Zealand to its former Prime Minister John Key (BRIO), claiming a fictional short story is a BBC documentary (SimCLS), or adding to a textbook excerpt on the Civil War by calling it the longest,  most expensive conflict in US history (BRIO and SimCLS). Below we provide a comparison of outputs for two documents and a qualitative analysis. 

We also asked evaluators whether they could tell if summaries were NLG outputs, and learned that while `NLG' guesses were correct, and most human summaries were also recognized, humans could not tell for certain in 56\% of the outputs they evaluated (incl.~8\% of human-written cases).

\begin{figure}[ht]
\centering
  \subfigure[\textit{conversation}]{\includegraphics[width=\columnwidth]{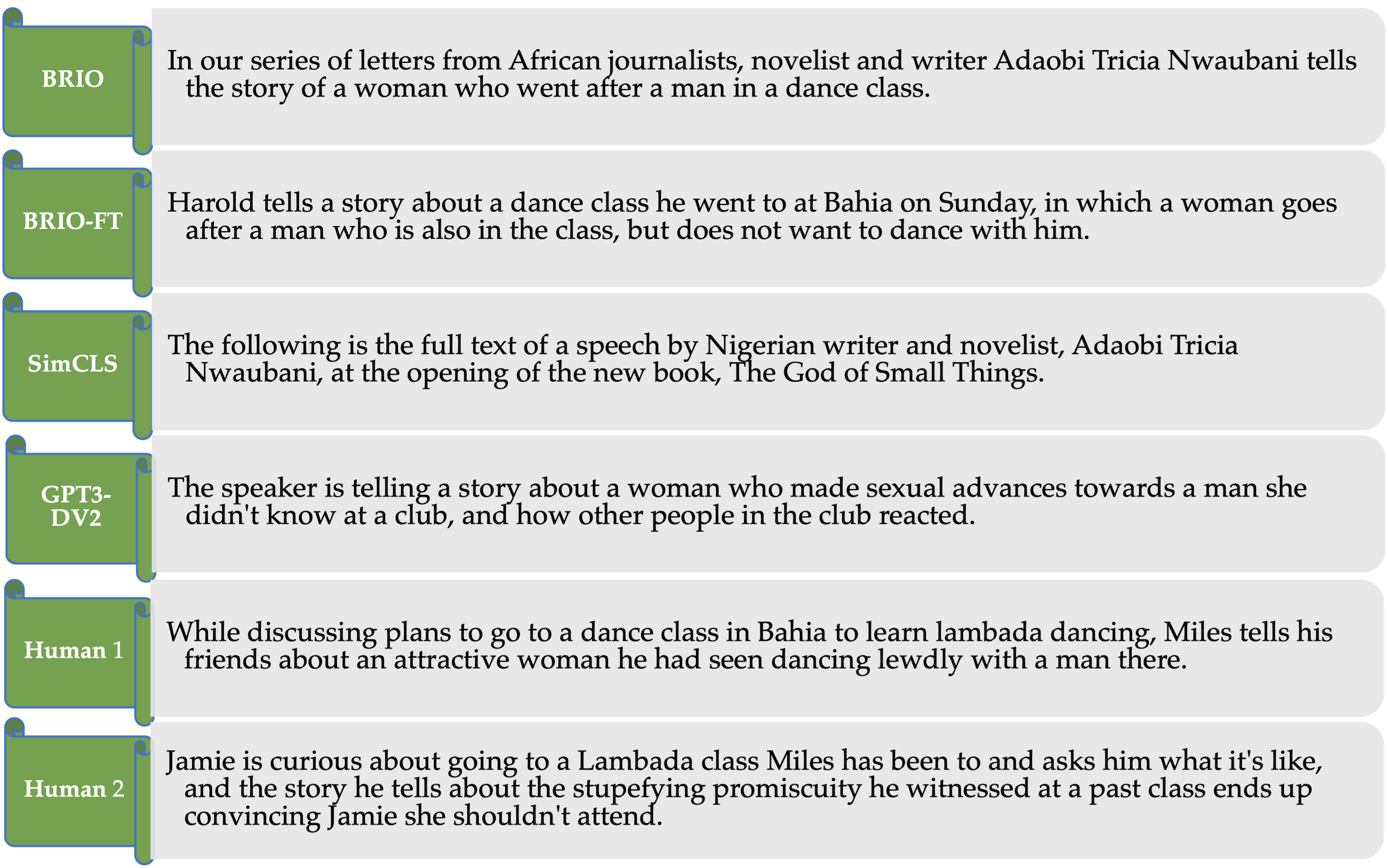}\label{fig:lambda}}
  \subfigure[\textit{news}]{\includegraphics[width=\columnwidth]{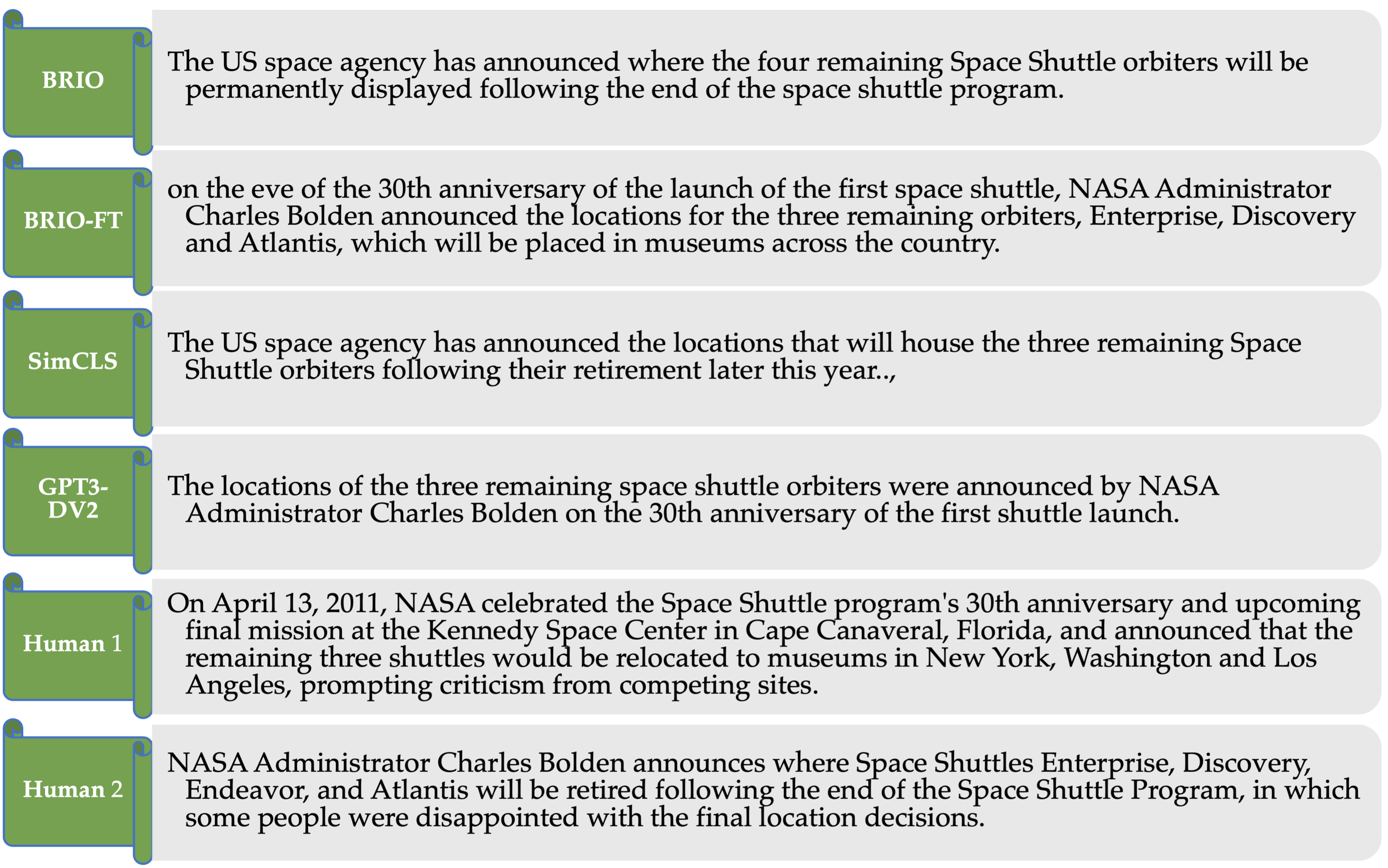}\label{fig:nasa}}
  \vspace{-8pt}
  \caption{Sample Summary Outputs of a \textit{conversation} Text and a \textit{news} Text from Each Evaluation Scenario. }
  \label{fig:qualit-analysis-examples}
  \vspace{-12pt}
\end{figure}

\paragraph{Qualitative Analysis}
Figure \ref{fig:qualit-analysis-examples} shows two human-written and several system-generated summaries, for a conversation in (a) and for a news text in (b).\footnote{The PDFs of the full-text of these two documents are provided in the repository of the paper for reference. } Note the typical hallucinated lead about journalists in the first BRIO output, which disappears after fine-tuning, and a similar insertion about a Nigerian writer in the output for SimCLS. GPT3-DV2 does not show similar issues, but misses important context information, e.g.~the purpose of the conversation revolving around whether speakers should go to a specific dance class, and why or why not. 

The news output is substantially better for all systems. BRIO disagrees with SimCLS and GPT3 on the number of `remaining' space shuttles: three remained to be retired, but there were four total in the article, including the already retired shuttle Discovery. All pre-trained system outputs are substantially less detailed than the human summaries, which add information about time and place of the announcement, or the list of space shuttles. Human 2 commits a similar hallucination error to BRIO in identifying the already retired Discovery as being retired at document creation time. However, both human summaries agree that a prominent part of the story involves the disappointment or criticism from sites that were not selected to house retired shuttles, a topic to which most of the latter half of the original story is dedicated. The fine-tuned model successfully adds more details in line with the human summaries, but also fails to capture the site controversy in the second half of the document.

\section{Conclusion}
\label{sec:conclusion}

% As the results from the previous two sections make clear, 
The dataset and guidelines introduced in this paper make a step towards consistent and constrained multi-genre evaluation of factual summarization. Our results show that domain-general summarization is still hampered by serious reliability and factuality problems, which may only become apparent when confronted with a dataset with strict `reality check' constraints and diverse text types. Even small amounts of such data can be used to fine-tune pre-trained systems, with measurable improvements for system outputs.

The human evaluation study also revealed that pre-trained systems are bad at delivering substitutive summaries, perhaps because, as pointed out in \citet{ehudreiter2022post}, ``summarisation datasets should contain summaries,'' but often they do not. Meanwhile, human identification of possibly more minor hallucinations in human-written summaries also suggests that more work is needed in delimiting what a `reality check' for summaries should include.

\section*{Limitations}
\label{sec:limitations}

GUMSum is designed to constrain summaries to one sentence for all 12 genres, which raises the question of whether one-sentence summaries are useful for all possible genres or long-document summarization. This is a complex topic that needs in-depth investigation. For GUMSum, as mentioned in Section \ref{sec:dataset}, document length is limited to 167--1,878 tokens. Moreover, in analyzing human evaluators' responses to two open-ended questions ([\ref{open-ended-question1}] and [\ref{open-ended-question2}] in Appendix \ref{appendix:humaneval-details}), we noticed that virtually all evaluators mentioned that limiting the summary to one-sentence is very difficult and that some genres were easier than others. For example, one evaluator who was given a vlog and a travel guide commented that, 

\begin{quote}
    ``\textit{The travel guide was much more difficult than the vlog, likely because it was longer and denser. [...] the travel guide packed a lot more information into its pages and within each sentence}.'' 
\end{quote}
\noindent This indicates that genre differences at the summary-level is not trivial due to the style of the original text.

Additionally, this paper examined a specific subset of pre-trained systems and one version of GPT3's pretrained language model (i.e.~\texttt{GPT3-text-davinci-002}), producing findings which may not generalize to other settings. The dataset used for the evaluation is also substantially smaller than those used in most work on summarization, due to the fact that it was carefully crafted based on both general and genre-specific guidelines to be substitutive and to avoid hallucinations and faithfulness issues, rather than originating in a found dataset, in order to conduct a more targeted evaluation, as recommended by \citet{RevisitingGoldStandard4Summarization}. While it is inevitable that more data would lead to different results, we do not believe that system rankings or overall findings would be substantially different, so long as the guidelines and genres examined here remain stable.

Finally, we must raise a further limitation involving text type and language: our study encompasses 12 specific written and spoken genres available in the UD English GUM corpus, but does not capture findings for other genres, or indeed other languages, which deserve more attention in future studies.

\section*{Ethics Statement}
\label{sec:ethics}

The data produced in this paper is made openly available in accordance with the original licenses of the underlying resources and academic fair use. we are keenly aware that NLP, and particularly NLG technology can be misused adversely, for example to generate fake news, we believe the risks posed by models which are not `reality-checked' outweigh those associated with improving models to prevent factuality and generalization issues across domains. The latter issue is particularly relevant, since technologies limited to particular domains and styles will primarily benefit actors in sectors engaged with that data (e.g.~news, for example, financial reporting), while underserving the public in other areas (e.g.~computer-mediated communication). We therefore concur with this year's ACL theme that work towards `reality checking' our outputs is a net positive.

\section*{Acknowledgements}
The human evaluation study was funded by a GSAS-GradGov Research Project Award (GRPA) towards graduate students' research and professional development endeavors at Georgetown University. 
We thank the following participants for their valuable participation and insightful feedback in our human evaluation study (alphabetically ordered by last names): Kris Cook, Jessica Cusi, Helen Dominic, Luke Gessler, Caroline Gish, Lauren Levine, Cynthia Li, Kristina Lignell, Davide Locatelli, Emma Manning, and others who prefer to stay anonymous. 
We thank Nathan Schneider and the anonymous reviewers for their feedback.

\bibliography{anthology,custom}

\begin{thebibliography}{32}
\expandafter\ifx\csname natexlab\endcsname\relax\def\natexlab#1{#1}\fi

\bibitem[{Banerjee and Lavie(2005)}]{banerjee-lavie-2005-meteor}
Satanjeev Banerjee and Alon Lavie. 2005.
\newblock \href {https://aclanthology.org/W05-0909} {{METEOR}: An automatic
  metric for {MT} evaluation with improved correlation with human judgments}.
\newblock In \emph{Proceedings of the {ACL} Workshop on Intrinsic and Extrinsic
  Evaluation Measures for Machine Translation and/or Summarization}, pages
  65--72, Ann Arbor, Michigan. Association for Computational Linguistics.

\bibitem[{Brown et~al.(2020)Brown, Mann, Ryder, Subbiah, Kaplan, Dhariwal,
  Neelakantan, Shyam, Sastry, Askell, Agarwal, Herbert-Voss, Krueger, Henighan,
  Child, Ramesh, Ziegler, Wu, Winter, Hesse, Chen, Sigler, Litwin, Gray, Chess,
  Clark, Berner, McCandlish, Radford, Sutskever, and
  Amodei}]{GPT3BrownEtAal2020}
Tom Brown, Benjamin Mann, Nick Ryder, Melanie Subbiah, Jared~D Kaplan, Prafulla
  Dhariwal, Arvind Neelakantan, Pranav Shyam, Girish Sastry, Amanda Askell,
  Sandhini Agarwal, Ariel Herbert-Voss, Gretchen Krueger, Tom Henighan, Rewon
  Child, Aditya Ramesh, Daniel Ziegler, Jeffrey Wu, Clemens Winter, Chris
  Hesse, Mark Chen, Eric Sigler, Mateusz Litwin, Scott Gray, Benjamin Chess,
  Jack Clark, Christopher Berner, Sam McCandlish, Alec Radford, Ilya Sutskever,
  and Dario Amodei. 2020.
\newblock \href
  {https://proceedings.neurips.cc/paper/2020/file/1457c0d6bfcb4967418bfb8ac142f64a-Paper.pdf}
  {{Language Models are Few-Shot Learners}}.
\newblock In \emph{Advances in Neural Information Processing Systems (NIPS)},
  volume~33 of \emph{NIPS'20}, pages 1877--1901, Red Hook, NY, USA. Curran
  Associates, Inc.

\bibitem[{Edmundson(1969)}]{Edmundson1969AutomaticExtracting}
H.~P. Edmundson. 1969.
\newblock \href {https://doi.org/10.1145/321510.321519} {{New Methods in
  Automatic Extracting}}.
\newblock \emph{J. ACM}, 16(2):264–285.

\bibitem[{Fabbri et~al.(2021)Fabbri, Kry{\'s}ci{\'n}ski, McCann, Xiong, Socher,
  and Radev}]{fabbri-etal-2021-summeval}
Alexander~R. Fabbri, Wojciech Kry{\'s}ci{\'n}ski, Bryan McCann, Caiming Xiong,
  Richard Socher, and Dragomir Radev. 2021.
\newblock \href {https://doi.org/10.1162/tacl_a_00373} {{S}umm{E}val:
  Re-evaluating summarization evaluation}.
\newblock \emph{Transactions of the Association for Computational Linguistics},
  9:391--409.

\bibitem[{Gehrmann et~al.(2022)Gehrmann, Clark, and
  Sellam}]{Gehrmann2022ObstaclesInNLGEval}
Sebastian Gehrmann, Elizabeth Clark, and Thibault Sellam. 2022.
\newblock \href {https://doi.org/10.48550/ARXIV.2202.06935} {Repairing the
  cracked foundation: A survey of obstacles in evaluation practices for
  generated text}.
\newblock \emph{arXiv}.

\bibitem[{Goyal et~al.(2022)Goyal, Li, and Durrett}]{NewsSumGPT3}
Tanya Goyal, Junyi~Jessy Li, and Greg Durrett. 2022.
\newblock \href {https://doi.org/10.48550/ARXIV.2209.12356} {News
  {S}ummarization and {E}valuation in the {E}ra of {GPT-3}}.
\newblock \emph{arXiv}.

\bibitem[{Hermann et~al.(2015)Hermann, Ko\v{c}isk\'{y}, Grefenstette, Espeholt,
  Kay, Suleyman, and Blunsom}]{CNNDM}
Karl~Moritz Hermann, Tom\'{a}\v{s} Ko\v{c}isk\'{y}, Edward Grefenstette, Lasse
  Espeholt, Will Kay, Mustafa Suleyman, and Phil Blunsom. 2015.
\newblock Teaching machines to read and comprehend.
\newblock In \emph{Proceedings of the 28th International Conference on Neural
  Information Processing Systems - Volume 1}, NIPS'15, page 1693–1701,
  Cambridge, MA, USA. MIT Press.

\bibitem[{Kim et~al.(2019)Kim, Kim, and Kim}]{kim-etal-2019-abstractive}
Byeongchang Kim, Hyunwoo Kim, and Gunhee Kim. 2019.
\newblock \href {https://doi.org/10.18653/v1/N19-1260} {Abstractive
  summarization of {R}eddit posts with multi-level memory networks}.
\newblock In \emph{Proceedings of the 2019 Conference of the North {A}merican
  Chapter of the Association for Computational Linguistics: Human Language
  Technologies, Volume 1 (Long and Short Papers)}, pages 2519--2531,
  Minneapolis, Minnesota. Association for Computational Linguistics.

\bibitem[{Lin(2004)}]{lin-2004-rouge}
Chin-Yew Lin. 2004.
\newblock \href {https://aclanthology.org/W04-1013} {{ROUGE}: A package for
  automatic evaluation of summaries}.
\newblock In \emph{Text Summarization Branches Out}, pages 74--81, Barcelona,
  Spain. Association for Computational Linguistics.

\bibitem[{Liu et~al.(2022{\natexlab{a}})Liu, Fabbri, Liu, Zhao, Nan, Han, Han,
  Joty, Wu, Xiong, and Radev}]{RevisitingGoldStandard4Summarization}
Yixin Liu, Alexander~R. Fabbri, Pengfei Liu, Yilun Zhao, Linyong Nan, Ruilin
  Han, Simeng Han, Shafiq Joty, Chien-Sheng Wu, Caiming Xiong, and Dragomir
  Radev. 2022{\natexlab{a}}.
\newblock \href {https://doi.org/10.48550/ARXIV.2212.07981} {Revisiting the
  gold standard: Grounding summarization evaluation with robust human
  evaluation}.
\newblock \emph{arXiv}.

\bibitem[{Liu and Liu(2021)}]{liu-liu-2021-simcls}
Yixin Liu and Pengfei Liu. 2021.
\newblock \href {https://doi.org/10.18653/v1/2021.acl-short.135} {{S}im{CLS}: A
  simple framework for contrastive learning of abstractive summarization}.
\newblock In \emph{Proceedings of the 59th Annual Meeting of the Association
  for Computational Linguistics and the 11th International Joint Conference on
  Natural Language Processing (Volume 2: Short Papers)}, pages 1065--1072,
  Online. Association for Computational Linguistics.

\bibitem[{Liu et~al.(2022{\natexlab{b}})Liu, Liu, Radev, and
  Neubig}]{liu-etal-2022-brio}
Yixin Liu, Pengfei Liu, Dragomir Radev, and Graham Neubig. 2022{\natexlab{b}}.
\newblock \href {https://doi.org/10.18653/v1/2022.acl-long.207} {{BRIO}:
  Bringing order to abstractive summarization}.
\newblock In \emph{Proceedings of the 60th Annual Meeting of the Association
  for Computational Linguistics (Volume 1: Long Papers)}, pages 2890--2903,
  Dublin, Ireland. Association for Computational Linguistics.

\bibitem[{Maddela et~al.(2022)Maddela, Kulkarni, and
  Preotiuc-Pietro}]{maddela-etal-2022-entsum}
Mounica Maddela, Mayank Kulkarni, and Daniel Preotiuc-Pietro. 2022.
\newblock \href {https://doi.org/10.18653/v1/2022.acl-long.237} {{E}nt{SUM}: A
  data set for entity-centric extractive summarization}.
\newblock In \emph{Proceedings of the 60th Annual Meeting of the Association
  for Computational Linguistics (Volume 1: Long Papers)}, pages 3355--3366,
  Dublin, Ireland. Association for Computational Linguistics.

\bibitem[{Marasović(2018)}]{MarasovicGradient2018NLP}
Ana Marasović. 2018.
\newblock \href
  {https://thegradient.pub/frontiers-of-generalization-in-natural-language-processing}
  {{NLP’s generalization problem, and how researchers are tackling it}}.
\newblock \emph{The Gradient}.

\bibitem[{Nan et~al.(2021)Nan, Nallapati, Wang, Nogueira~dos Santos, Zhu,
  Zhang, McKeown, and Xiang}]{nan-etal-2021-entity}
Feng Nan, Ramesh Nallapati, Zhiguo Wang, Cicero Nogueira~dos Santos, Henghui
  Zhu, Dejiao Zhang, Kathleen McKeown, and Bing Xiang. 2021.
\newblock \href {https://doi.org/10.18653/v1/2021.eacl-main.235} {Entity-level
  factual consistency of abstractive text summarization}.
\newblock In \emph{Proceedings of the 16th Conference of the European Chapter
  of the Association for Computational Linguistics: Main Volume}, pages
  2727--2733, Online. Association for Computational Linguistics.

\bibitem[{Narayan et~al.(2018)Narayan, Cohen, and
  Lapata}]{narayan-etal-2018-dont}
Shashi Narayan, Shay~B. Cohen, and Mirella Lapata. 2018.
\newblock \href {https://doi.org/10.18653/v1/D18-1206} {Don{'}t give me the
  details, just the summary! topic-aware convolutional neural networks for
  extreme summarization}.
\newblock In \emph{Proceedings of the 2018 Conference on Empirical Methods in
  Natural Language Processing}, pages 1797--1807, Brussels, Belgium.
  Association for Computational Linguistics.

\bibitem[{Nenkova and McKeown(2011)}]{Nenkova2011McKeown}
Ani Nenkova and Kathleen~R. McKeown. 2011.
\newblock \href {https://doi.org/10.1561/1500000015} {{Automatic
  Summarization}}.
\newblock \emph{Foundations and Trends in Information Retrieval},
  5(2-3):103--233.

\bibitem[{Novikova et~al.(2017)Novikova, Du{\v{s}}ek, Cercas~Curry, and
  Rieser}]{novikova-etal-2017-need}
Jekaterina Novikova, Ond{\v{r}}ej Du{\v{s}}ek, Amanda Cercas~Curry, and Verena
  Rieser. 2017.
\newblock \href {https://doi.org/10.18653/v1/D17-1238} {Why we need new
  evaluation metrics for {NLG}}.
\newblock In \emph{Proceedings of the 2017 Conference on Empirical Methods in
  Natural Language Processing}, pages 2241--2252, Copenhagen, Denmark.
  Association for Computational Linguistics.

\bibitem[{Papineni et~al.(2002)Papineni, Roukos, Ward, and Zhu}]{BLEU2002}
Kishore Papineni, Salim Roukos, Todd Ward, and Wei-Jing Zhu. 2002.
\newblock \href {https://doi.org/10.3115/1073083.1073135} {{BLEU: A Method for
  Automatic Evaluation of Machine Translation}}.
\newblock In \emph{Proceedings of the 40th Annual Meeting on Association for
  Computational Linguistics}, ACL '02, page 311–318, USA. Association for
  Computational Linguistics.

\bibitem[{Reiter(2018)}]{reiter2018structured}
Ehud Reiter. 2018.
\newblock \href {https://doi.org/10.1162/coli_a_00322} {{A Structured Review of
  the Validity of BLEU}}.
\newblock \emph{Computational Linguistics}, 44(3):393--401.

\bibitem[{Reiter(2022)}]{ehudreiter2022post}
Ehud Reiter. 2022.
\newblock \href {https://ehudreiter.com/2022/10/13/summarisation-datasets/}
  {Summarisation datasets should contain summaries!}

\bibitem[{Rezapour et~al.(2022)Rezapour, Reddy, Jones, and
  Soboroff}]{PodcastSummary2022SIGIR}
Rezvaneh Rezapour, Sravana Reddy, Rosie Jones, and Ian Soboroff. 2022.
\newblock \href {https://doi.org/10.1145/3477495.3531802} {What makes a good
  podcast summary?}
\newblock In \emph{Proceedings of the 45th International ACM SIGIR Conference
  on Research and Development in Information Retrieval}, SIGIR '22, page
  2039–2046, New York, NY, USA. Association for Computing Machinery.

\bibitem[{Sellam et~al.(2020)Sellam, Das, and Parikh}]{sellam-etal-2020-bleurt}
Thibault Sellam, Dipanjan Das, and Ankur Parikh. 2020.
\newblock \href {https://doi.org/10.18653/v1/2020.acl-main.704} {{BLEURT}:
  Learning robust metrics for text generation}.
\newblock In \emph{Proceedings of the 58th Annual Meeting of the Association
  for Computational Linguistics}, pages 7881--7892, Online. Association for
  Computational Linguistics.

\bibitem[{Tang et~al.(2022)Tang, Goyal, Fabbri, Laban, Xu, Yahvuz,
  Kryściński, Rousseau, and
  Durrett}]{UnderstandingFactualErrorsinSum2022Durrett}
Liyan Tang, Tanya Goyal, Alexander~R. Fabbri, Philippe Laban, Jiacheng Xu,
  Semih Yahvuz, Wojciech Kryściński, Justin~F. Rousseau, and Greg Durrett.
  2022.
\newblock \href {https://doi.org/10.48550/ARXIV.2205.12854} {{Understanding
  Factual Errors in Summarization: Errors, Summarizers, Datasets, Error
  Detectors}}.
\newblock \emph{arXiv}.

\bibitem[{Thomson et~al.(2023)Thomson, Reiter, and
  Sundararajan}]{THOMSON2023101482}
Craig Thomson, Ehud Reiter, and Barkavi Sundararajan. 2023.
\newblock \href {https://doi.org/https://doi.org/10.1016/j.csl.2023.101482}
  {Evaluating factual accuracy in complex data-to-text}.
\newblock \emph{Computer Speech \& Language}.

\bibitem[{Wolf et~al.(2019)Wolf, Debut, Sanh, Chaumond, Delangue, Moi, Cistac,
  Rault, Louf, Funtowicz, Davison, Shleifer, von Platen, Ma, Jernite, Plu, Xu,
  Scao, Gugger, Drame, Lhoest, and Rush}]{HF_Transformers}
Thomas Wolf, Lysandre Debut, Victor Sanh, Julien Chaumond, Clement Delangue,
  Anthony Moi, Pierric Cistac, Tim Rault, Rémi Louf, Morgan Funtowicz, Joe
  Davison, Sam Shleifer, Patrick von Platen, Clara Ma, Yacine Jernite, Julien
  Plu, Canwen Xu, Teven~Le Scao, Sylvain Gugger, Mariama Drame, Quentin Lhoest,
  and Alexander~M. Rush. 2019.
\newblock \href {https://doi.org/10.48550/ARXIV.1910.03771} {{HuggingFace's
  Transformers: State-of-the-art Natural Language Processing}}.
\newblock \emph{arXiv}.

\bibitem[{Xiao and Carenini(2022)}]{XiaoCarenini2022}
Wen Xiao and Giuseppe Carenini. 2022.
\newblock \href {https://arxiv.org/abs/2209.03479} {{Entity-based SpanCopy for
  Abstractive Summarization to Improve the Factual Consistency}}.
\newblock \emph{Arxiv Preprint}.

\bibitem[{Zeldes(2017)}]{Zeldes2017}
Amir Zeldes. 2017.
\newblock \href {https://doi.org/http://dx.doi.org/10.1007/s10579-016-9343-x}
  {The {GUM} {C}orpus: {Creating Multilayer Resources in the Classroom}}.
\newblock \emph{Language Resources and Evaluation}, 51(3):581--612.

\bibitem[{Zhang et~al.(2020)Zhang, Kishore, Wu*, Weinberger, and
  Artzi}]{Zhang2020BERTScore}
Tianyi Zhang, Varsha Kishore, Felix Wu*, Kilian~Q. Weinberger, and Yoav Artzi.
  2020.
\newblock \href {https://openreview.net/forum?id=SkeHuCVFDr} {{BERTScore:
  Evaluating Text Generation with BERT}}.
\newblock In \emph{International Conference on Learning Representations}.

\bibitem[{Zhao et~al.(2019)Zhao, Peyrard, Liu, Gao, Meyer, and
  Eger}]{zhao-etal-2019-moverscore}
Wei Zhao, Maxime Peyrard, Fei Liu, Yang Gao, Christian~M. Meyer, and Steffen
  Eger. 2019.
\newblock \href {https://doi.org/10.18653/v1/D19-1053} {{M}over{S}core: Text
  generation evaluating with contextualized embeddings and earth mover
  distance}.
\newblock In \emph{Proceedings of the 2019 Conference on Empirical Methods in
  Natural Language Processing and the 9th International Joint Conference on
  Natural Language Processing (EMNLP-IJCNLP)}, pages 563--578, Hong Kong,
  China. Association for Computational Linguistics.

\bibitem[{Zhu et~al.(2021)Zhu, Hinthorn, Xu, Zeng, Zeng, Huang, and
  Jiang}]{zhu-etal-2021-enhancing}
Chenguang Zhu, William Hinthorn, Ruochen Xu, Qingkai Zeng, Michael Zeng,
  Xuedong Huang, and Meng Jiang. 2021.
\newblock \href {https://doi.org/10.18653/v1/2021.naacl-main.58} {Enhancing
  factual consistency of abstractive summarization}.
\newblock In \emph{Proceedings of the 2021 Conference of the North American
  Chapter of the Association for Computational Linguistics: Human Language
  Technologies}, pages 718--733, Online. Association for Computational
  Linguistics.

\bibitem[{Zopf et~al.(2016)Zopf, Peyrard, and
  Eckle-Kohler}]{zopf-etal-2016-next}
Markus Zopf, Maxime Peyrard, and Judith Eckle-Kohler. 2016.
\newblock \href {https://aclanthology.org/C16-1145} {The next step for
  multi-document summarization: A heterogeneous multi-genre corpus built with a
  novel construction approach}.
\newblock In \emph{Proceedings of {COLING} 2016, the 26th International
  Conference on Computational Linguistics: Technical Papers}, pages 1535--1545,
  Osaka, Japan. The COLING 2016 Organizing Committee.

\end{thebibliography}
\bibliographystyle{acl_natbib}

\appendix

\section{Genre-specific Guidelines}
\label{appendix:guidelines}

The following excerpts from genre-specific guidelines exemplify instructions which were given to annotators working on documents in those specific genres. The full guidelines can be viewed at  \url{https://wiki.gucorpling.org/gum/summarization}.

\subsection{Biographies}

Summaries for biographies and other texts centered around an individual:

\begin{itemize}
    \item typically take the form ``Kim is/was a French X who ... ''
    \item typically include information about what this person is/was known for (``... best known for ...'')
    \item information about the time period and place is typically included (``a Japanese X'', ``a German X living in France'', ``a 19th century Kenyan X'')
\end{itemize}

\noindent Examples:

\begin{itemize}
    \item Jared Padalecki is an award winning American actor who gained prominence in the series Gilmore Girls, best known for playing the role of Sam Winchester in the TV series Supernatural, and for his active role in campaigns to support people struggling with depression, addiction, suicide and self-harm.
    \item Jenna Nicole Mourey, better known as Jenna Marbles, is a very successful American YouTube personality, vlogger, comedian and actress, known for her videos "How To Trick People Into Thinking You're Good Looking" and "How To Avoid Talking To People You Don't Want To Talk To".
\end{itemize}

\subsection{Fiction}

\begin{itemize}
    \item  In non-metalinguistic texts (i.e.~fiction itself, not texts about fiction), summarize the text as if it is a literal, true story; for example, ``Huckleberry Finn is fishing'', not ``In this extract from the novel Huckleberry Finn, fictional character Huck is...''
    \item Even if described events are factually incorrect, or involve science fiction or imaginary contexts, we summarize without commenting on this (e.g.~``Three unicorns chat and decide to go fishing'')
    \item Unnamed active protagonists should be referred to as ``a/the protagonist''
    \item An unnamed narrator who is not an agent in the story can be referred to as ``a/the narrator''

\end{itemize}

\noindent  Examples:

\begin{itemize}
    \item Jacques Chalmers, a starfighter pilot for the Empire, is terrified of overwhelming enemy forces as he leaves his deployment carrier together with his comrades, and later narrowly escapes the Enemy after witnessing the destruction of the Kethlan system.
    \item Santa Claus's second wife, Betty Moroz, plays online video games with her friends Williams and Gomez while making dinner on Christmas Eve, and is then disappointed when Santa gets a call from his secretary Ginny and goes out to take care of the children of the world, missing dinner.
\end{itemize}

\subsection{Vlogs}

\begin{itemize}
    \item Typically a present tense third person style is used, and events are ordered in sequence, for example: ``Ash tells about her day, which includes a yoga class, marketing brand management class, doing some work while having coffee at Saxby's, and finally cooking pasta with peppers for dinner together with her boyfriend Harry.''
    \item As in conversations, people other than the vlogger who play a significant role in the vlog should be mentioned, but if their name is not mentioned within the excerpt being annotated, then they can only be referred to using generic terms (``a friend/relative/...'')
    \item If the vlogger does not mention that they are a vlogger in the video, or that this is a vlog, do not refer to them as such (e.g.~``Jasmine tells about ...'', not ``YoutTube vlogger Jasmine tells ...'')
\end{itemize}

\noindent Examples:

\begin{itemize}
    \item Jasmine tells about how she tested positive for Covid on December 16th after she spent time without a mask with her sister, who also tested positive, and recounts her symptoms over several days, starting from a sore throat, then fever and congestion, and finally a partial loss of smell and taste and shortness of breath.

\end{itemize}

\section{Experiment Details}
\label{appendix:experiment-details}

\subsection{Fine-tuning on BRIO}

All three fine-tuning sessions were conducted using 1 NVIDIA A100 40GB GPU on Google Cloud Platform, which cost \$2.8 per hour.\footnote{\url{https://cloud.google.com/compute/docs/gpus\#a100-40gb}} The configurations of BRIO for XSum\footnote{\url{https://github.com/yixinL7/BRIO/blob/main/config.py\#L37-L71}} were used except that the default number of \texttt{epochs} was increased to 1000 from 100 in order to achieve better validation performance on GUMSum's \texttt{dev} set. Specifically, we take BRIO's \textit{generation} model checkpoint on XSum from Huggingface's Transformers \cite{HF_Transformers}.\footnote{\url{https://huggingface.co/Yale-LILY/brio-xsum-cased}} The average training time for a single run was about 7 hours. Table \ref{tab:brio-ft-val} shows the validation performance of each run on the documents from the \texttt{dev} set of GUM V9. 
Both \texttt{dev} and \texttt{test} partitions contain 24 documents, 2 for each genre, leaving 165 documents for training.\footnote{The complete list of \texttt{train}/\texttt{dev}/\texttt{test} document names is provided in the repository.} 

\begin{table}[ht]
\centering
\resizebox{\columnwidth}{!}{%
\begin{tabular}{@{}lccccc@{}}
\toprule
 & \multicolumn{1}{l}{\textbf{\textsc{val}\_\textsc{loss}}} & \multicolumn{1}{l}{\textbf{\textsc{val}\_R-1}} & \multicolumn{1}{l}{\textbf{\textsc{val}\_R-2}} & \multicolumn{1}{l}{\textbf{\textsc{val}\_R-L}} & 
 \multicolumn{1}{l}{\textbf{\textsc{best}\_\textbf{epoch}}} \\ \midrule
\multicolumn{1}{l|}{RUN 1} & 72.3 & 39.3 & 14.5 & 29.3 & 899 \\
\multicolumn{1}{l|}{RUN 2} & 71.9 & 39.9 & 15.3 & 29.2 & 799 \\
\multicolumn{1}{l|}{RUN 3} & 73.0 & 38.3 & 14.1 & 28.6 & 849 \\ \midrule
\multicolumn{1}{c|}{\textsc{avg.}} & 72.4 & 39.1 & 14.6 & 29.0 & $-$ \\ \bottomrule
\end{tabular}%
}
\caption{FT Validation Performance on 24 \texttt{dev} docs. }
\label{tab:brio-ft-val}
\vspace{-10pt}
\end{table}

\subsection{GPT3 Output Selection}

We use OpenAI's \texttt{GPT3-text-davinci-002}\footnote{\texttt{GPT3-text-davinci-003} was not available at the time.} with the prompt \textit{Summarize the text above in one sentence.}~and keep the default settings.
% \footnote{\url{https://beta.openai.com/examples/default-tldr-summary}}
Due to the nondeterministic nature and in order to ensure a fair comparison, we generated 3 summaries for each text and computed average ROUGE scores (the mean of R-1/2/L) against the human-written summaries and selected the summary with the middle average ROUGE score. At the time, the Davinci model costs \$0.0200 / 1K tokens. To avoid repetitive computation and to facilitate further research, we release all the GPT3-generated summaries for GUMSum. No post-editing was made on the GPT3-generated summaries.

\subsection{BRIO-/SimCLS- Generated Summaries}

We use BRIO's \textit{generation} model checkpoint on XSum available on Huggingface (i.e.~\texttt{Yale-LILY/brio-xsum-cased}) to obtain BRIO-generated summaries for GUMSum's texts. For SimCLS \cite{liu-liu-2021-simcls}, we use the checkpoint on XSum provided by the authors in their GitHub repository.\footnote{\url{https://github.com/yixinL7/SimCLS}} Although some BRIO-/SimCLS-generated summaries contain trailing punctuation, no post-editing was made on these system outputs.

\section{Human Evaluation Details}
\label{appendix:humaneval-details}

We recruited 12 students who are native speakers of English to participate in this human evaluation study. Each student was assigned two documents from two different genres. 
They were given 4 weeks to work on a series of tasks for each document, as shown in Figure \ref{fig:humaneval1} below. 
Every student received a Google Form for each assigned text. 

\begin{figure}[ht]
    \centering
    \includegraphics[width=\columnwidth]{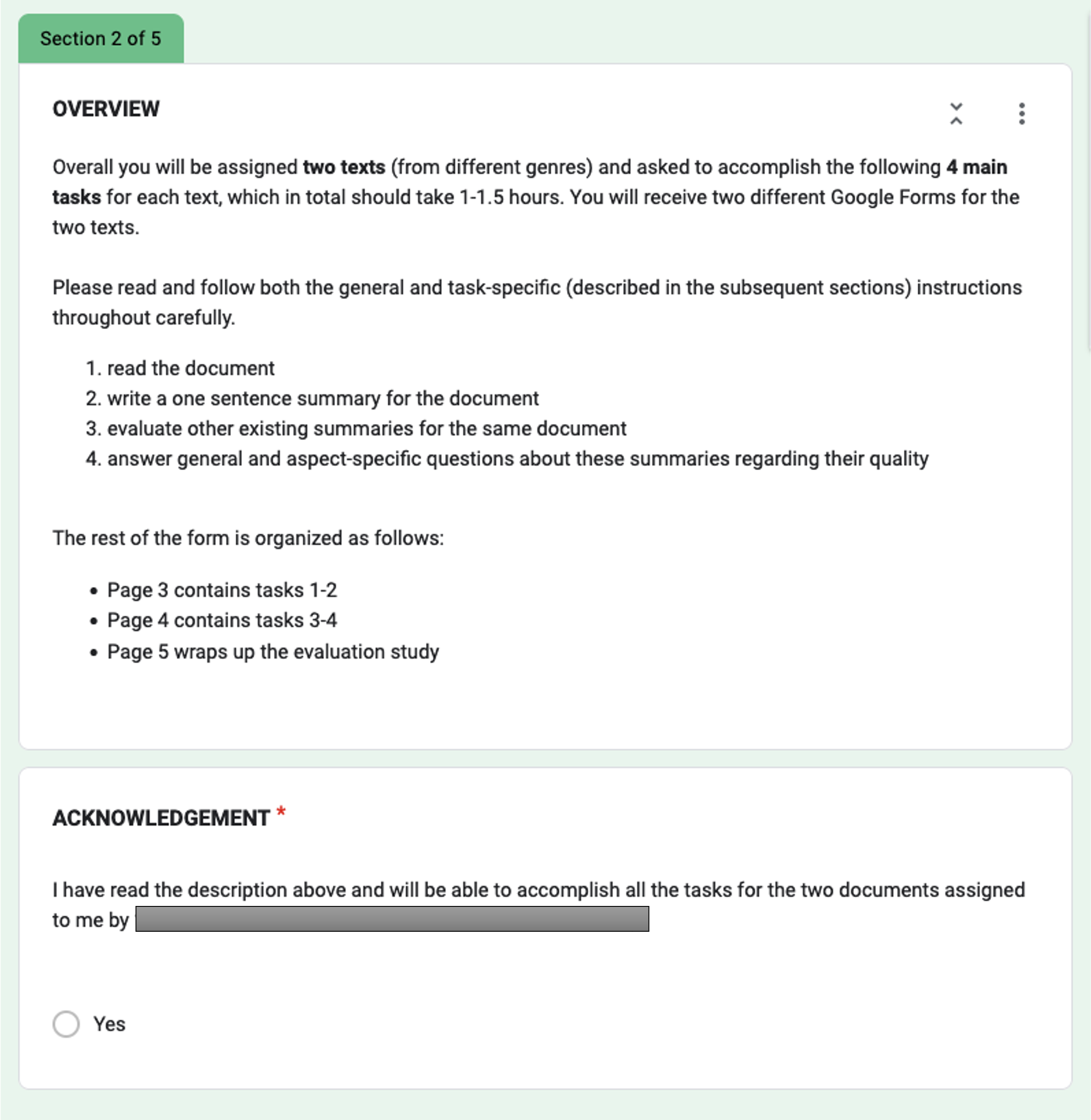}
    \caption{Overview of Task Description. }
    \label{fig:humaneval1}
    \vspace{-8pt}
\end{figure}

\begin{figure}[htp]
    \centering
    \includegraphics[width=\columnwidth]{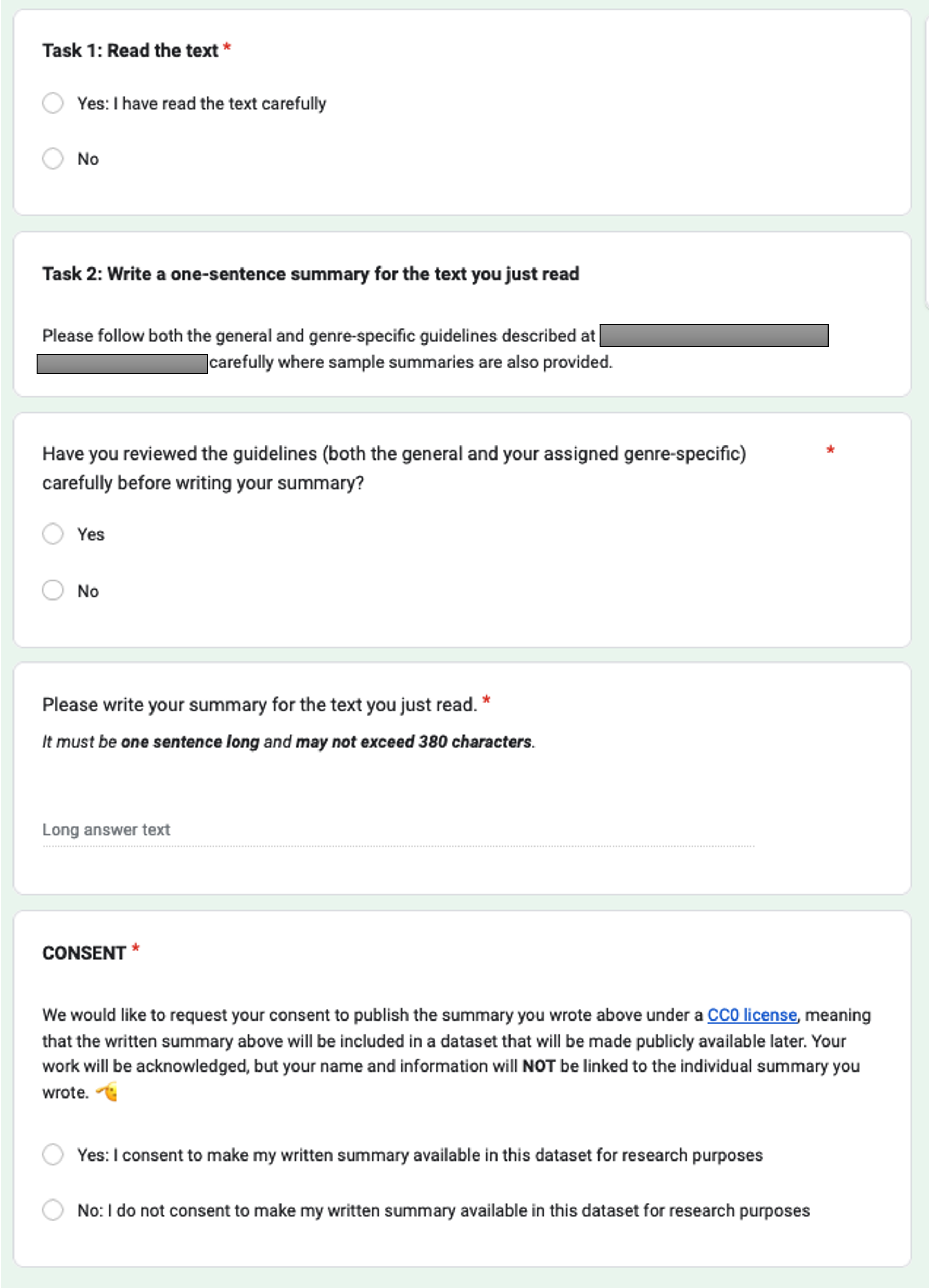}
    \caption{A Screenshot of Tasks 1 and 2. }
    \label{fig:humaneval2}
    \vspace{-8pt}
\end{figure}

\paragraph{Tasks 1 and 2} Students were asked to review both general and genre-specific guidelines before writing their own one-sentence summary for the assigned document. We also asked for their consent to release their written-summaries to GUMSum to facilitate multiple-reference evaluation and inter-annotator agreement, as shown in Figure \ref{fig:humaneval2}. 

\paragraph{Tasks 3 and 4} Students were presented both system-generated and human-written summaries in order to evaluate various aspects of each summary candidate. The order of outputs shown to the evaluators was randomized for each source text, and we also ask them to not modify their written summary after viewing the presented ones. In addition, we ask the evaluators to justify their decisions in a few sentences for certain questions: 

\begin{enumerate}
    \item \textbf{Please choose your most and least preferred summaries respectively.} You can select more than one for each category below if multiple summaries are equally most or least preferred by you. 
    \begin{itemize}
        \item Please justify your decisions above in a few sentences below. \textit{For instance, you could say,  "I prefer summary X over summary Y because X doesn't contain the main point (while a minor one is included) or Y contains incorrect information" etc. The more detailed the justifications, the better! }
    \end{itemize}
    \item \textbf{How substitutive is each summary candidate?} According to the guidelines, substitutive summaries replace reading the text as best as possible in one sentence - they are not just meant to attract readers to reading the text; they are meant to save you the trouble of reading it) 
    \item Does the summary include information \textbf{NOT PRESENT} in the text \textbf{even if you happen to know that it is factually correct}? 
    \begin{itemize}
        \item Please justify your decisions (esp.~the ones you chose YES for) above in a few sentences below. For instance, you can list the relevant information below. 
    \end{itemize}
    \item Does the summary include \textbf{INCORRECT information}? (i.e.~information \textbf{PRESENT} in the original text but used or interpreted \textbf{\textit{in a different, misleading, or incorrect way in the summary}}; in other words, this summary is not faithful to the original text)
    \begin{itemize}
        \item Please justify your decisions (esp. the ones you chose YES for) above in a few sentences below. For instance, you can list the relevant information below. 
    \end{itemize}
    \item \textbf{Is the summary written in good English? }(e.g.~no grammar errors or incomplete sentences etc.) 
    \item \textbf{Can you tell which summary is human-written and which one is computer-generated?} If you are very unsure about this (confidence level at or below 50\%), then choose the "can't tell" category. 
    \begin{itemize}
        \item Please justify your decisions above in a few sentences below. \textit{In particular, if you have a very strong opinion about a specific summary or certain summaries, we'd highly appreciate it if you could share your valuable thoughts with us. }
    \end{itemize}
\end{enumerate}

\paragraph{Wrapping-up} The last part of the evaluation study is to ask evaluators to first rate the level of difficulty of the entire evaluation task on a scale of 1 to 5 where 1 means `Not difficult at all' and 5 means `Extremely difficult'. We also collect their responses to the following open-ended questions in order to help us get a better idea of the challenges of producing a good summary for various text types, which are very valuable insights to guide future research on designing more specifically defined guidelines and targeted evaluation. 

\begin{enumerate}
    \item Based on your experience here, what's the most difficult or challenging thing you found when writing a one-sentence summary for the genre you are assigned? \label{open-ended-question1}
    \item Is there anything else you would like to share regarding your experience of writing a summary and/or evaluating other existing summaries? \label{open-ended-question2}
\end{enumerate}

\subsection{Additional Plots of Responses from the Human Evaluation Study}
\begin{figure}[ht]
\centering
  \subfigure[English Quality]{\includegraphics[width=\columnwidth]{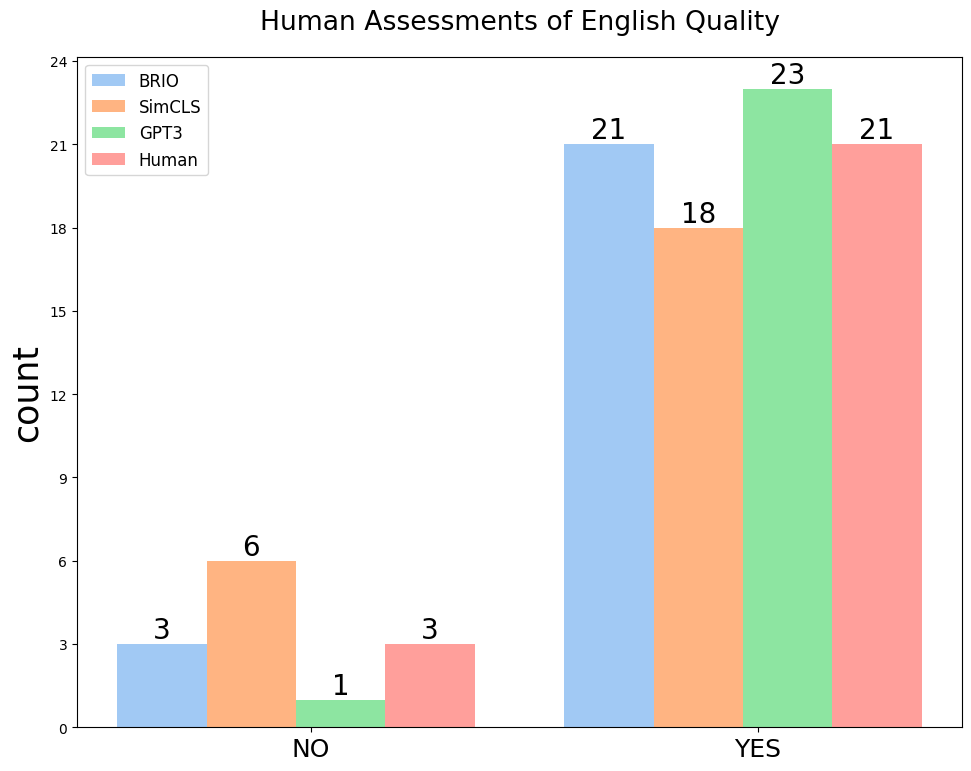}\label{fig:english-quality}}
  \subfigure[Source of Summaries]{\includegraphics[width=\columnwidth]{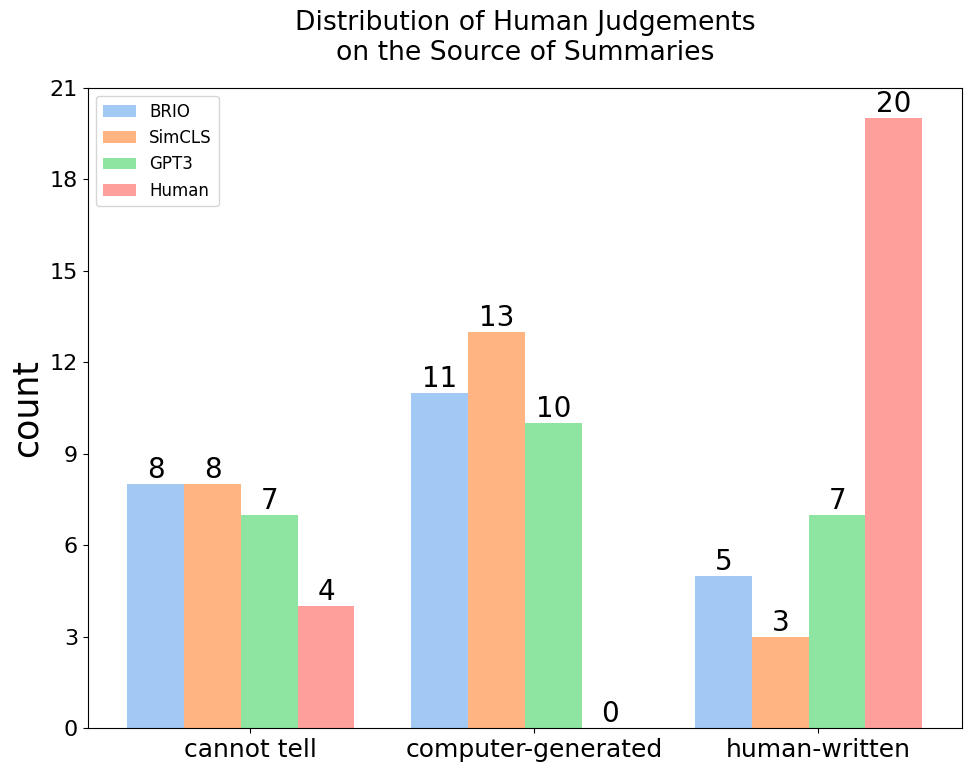}\label{fig:source}}
  \caption{Barplots of Human Evaluations on English Quality and Source of Summaries. }
\vspace{-8pt}
\label{fig:bar-plots-source-and-english}
\end{figure}

Figure \ref{fig:bar-plots-source-and-english} shows additional responses on English fluency quality for selected systems vs.~human performance, as well as a breakdown of annotators' guesses as to whether they were looking at human or system summaries. 

\end{document}